\documentclass[10pt,twocolumn,letterpaper]{article}

\usepackage{iccv}

\usepackage{times}
\usepackage{epsfig}
\usepackage{graphicx}
\usepackage{amsmath}
\usepackage{amssymb}
\usepackage{array}
\usepackage{cases}
\usepackage{color}
\usepackage{float}
\usepackage{multirow}
\usepackage{setspace}
\usepackage{txfonts}
\usepackage{algpseudocode}
\usepackage{booktabs} 
\usepackage{subfigure}
\usepackage{caption}
\usepackage{algorithm}
\usepackage{algpseudocode}
\usepackage{soul}
\usepackage{pifont}
\usepackage{times}
\usepackage{epsfig}
\usepackage{graphicx}
\usepackage{amssymb}
\usepackage{subfigure}
\usepackage{pifont}
\usepackage{color,xcolor,colortbl}
\usepackage{array}
\usepackage{url}
\usepackage{CJK}
\usepackage{makecell}
 \usepackage{arydshln}
\usepackage{array}     
\usepackage{longtable} 
\usepackage{colortab}  
\usepackage{dcolumn}
\usepackage{array}
\usepackage{setspace}
\usepackage{diagbox}
\usepackage{algorithmicx} 

\newcolumntype{I}{!{\vrule width 1.2pt}}
\newlength\savedwidth

\usepackage[pagebackref=true,breaklinks=true,colorlinks,bookmarks=false]{hyperref}

\iccvfinalcopy 

\ificcvfinal\fi

\begin{document}

\title{TransCrowd: weakly-supervised crowd counting with transformers}

\author{Dingkang Liang$^{1}$, Xiwu Chen$^{1}$, Wei Xu$^{2}$, Yu Zhou$^{1}$, Xiang Bai$^{1}$  \\
        $^{1}$Huazhong University of Science and Technology\\
        $^{2}$Beijing University of Posts and Telecommunications\\
}

\maketitle
\ificcvfinal\thispagestyle{empty}\fi

\begin{abstract}
The mainstream crowd counting methods usually utilize the convolution neural network (CNN) to regress a density map, requiring point-level annotations. However, annotating each person with a point is an expensive and laborious process. During the testing phase, the point-level annotations are not considered to evaluate the counting accuracy, which means the point-level annotations are redundant. Hence, it is desirable to develop weakly-supervised counting methods that just rely on count-level annotations, a more economical way of labeling. Current weakly-supervised counting methods adopt the CNN to regress a total count of the crowd by an image-to-count paradigm. However, having limited receptive ﬁelds for context modeling is an intrinsic limitation of these weakly-supervised CNN-based methods. These methods thus can not achieve satisfactory performance, with limited applications in the real-word. The Transformer is a popular sequence-to-sequence prediction model in NLP, which contains a global receptive field. In this paper, we propose TransCrowd, which reformulates the weakly-supervised crowd counting problem from the perspective of sequence-to-count based on Transformer. We observe that the proposed TransCrowd can effectively extract the semantic crowd information by using the self-attention mechanism of Transformer. To the best of our knowledge, this is the first work to adopt a pure Transformer for crowd counting research. Experiments on five benchmark datasets demonstrate that the proposed TransCrowd achieves superior performance compared with all the weakly-supervised CNN-based counting methods and gains highly competitive counting performance compared with some popular fully-supervised counting methods. An implementation of our method is available at~\url{https://github.com/dk-liang/TransCrowd}.
\end{abstract}


\maketitle

\section{Introduction}

Crowd counting is a hot topic in the computer vision community, which plays an essential role in video surveillance, public safety, and crowd analysis. Typical crowd counting methods~\cite{zhang2016single, li2018csrnet, xu2019learn,liu2021visdrone, liu2021visdrone} usually utilize the convolution neural network (CNN) to regress a density map, which has achieved significant progress recently. A standard regressor consists of an encoder and decoder: the encoder extracts the high-level feature information, and the decoder is designed for pixel-level regression based on the extracted feature. 

\begin{figure}[t]
	\begin{center}
		\includegraphics[width=0.86\linewidth]{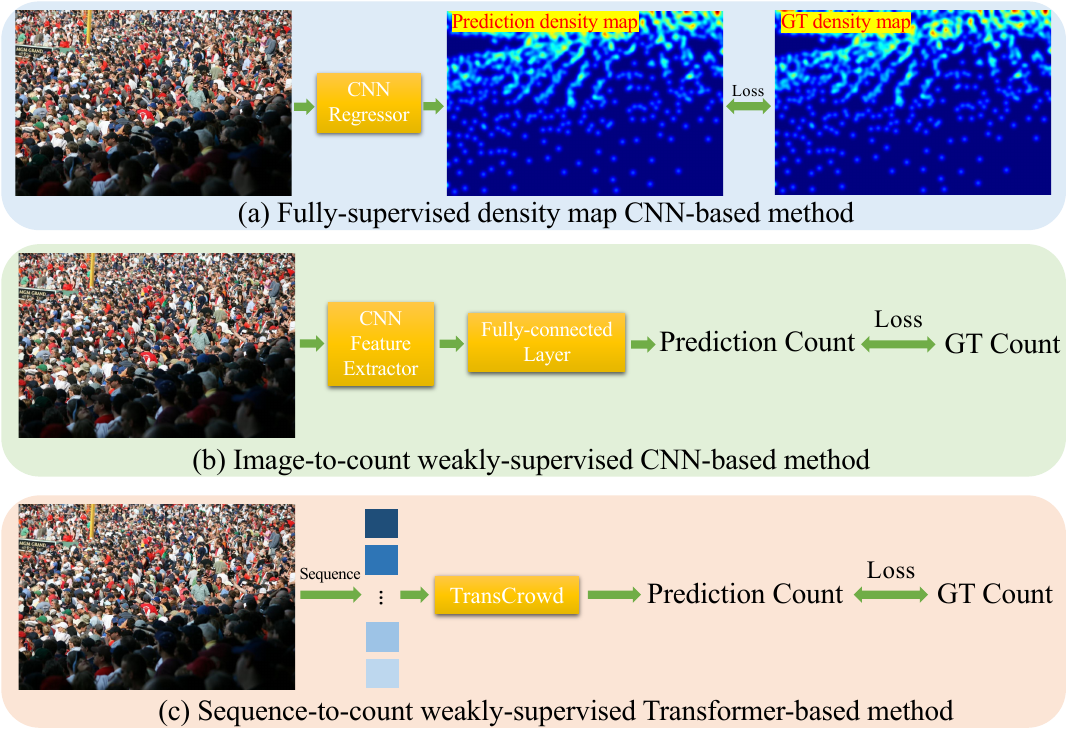}
	\end{center}
	\caption{(a) Traditional fully-supervised CNN-based method. All the training images are labeled with point-level annotations. (b) Weakly-supervised CNN-based method from image-to-count perspective, only relying on the annotated total count of the crowd. (c) The proposed TransCrowd, a weakly-supervised method,  reformulates the counting problem from the sequence-to-count perspective.}
	\label{fig:intro1}
	
\end{figure}

However, these density-map regression-based methods~\cite{zhang2016single, li2018csrnet, xu2019learn, bai2020adaptive} still have some drawbacks. 
1) They apply the point-level annotations to generate ground truth density maps, which are usually expensive cost. 
Actually, some methods~\cite{yang2020weakly, guo2015mobile, sheng2014leveraging, lei2021towards} discover that we can collect a new crowd dataset by using a more economical strategy, such as mobile crowd-sensing~\cite{guo2015mobile} technology or GPS-less~\cite{sheng2014leveraging} energy-efficient sensing scheduling. For a given crowd scene with different viewpoints and the total count keeps the same (such as auditoria, classroom), if we know the total count of one viewpoint, then the total count of other viewpoints is known. Besides, we can obtain the crowd number at a glance for some sparse crowd scenes.
(2) The annotated point label will not be taken to evaluating the counting performance, meaning the point label is redundant to some extent. Thus, point-level annotations are not absolutely necessary for the crowd counting task. 

Based on the above observations, it is desirable to develop the count-level crowd counting method. Following previous works~\cite{lei2021towards, yang2020weakly}, we call the methods which rely on the point-level annotations are fully-supervised paradigm, and the methods which only rely on count-level are weakly-supervised paradigm. The fully-supervised methods first utilize the point annotation to generate the ground truth density map and then elaborately design a regressor to generate a prediction density map and finally apply the $L2$ loss to measure the difference between the prediction and the ground truth, as shown in Fig.~\ref{fig:intro1}(a). The existing weakly-supervised methods usually regress the total count of crowd image directly, which is from the image-to-count perspective, as shown in Fig.~\ref{fig:intro1}(b).  

Recently, Transformer~\cite{vaswani2017attention}, a popular language model proposed by Google, has been explored in many vision tasks, such as classification~\cite{vit}, detection~\cite{detr, deformable_detr}, and segmentation~\cite{SETR}.
Unlike the CNN, which utilizes a limited receptive field, the Transformer~\cite{vaswani2017attention} provides the global receptive field, showing excellent advantages over pure CNN architectures. In this paper, we propose TransCrowd, which is the first to explore the Transformer into the weakly-supervised crowd counting task, establishing the perspective of sequence-to-count prediction, as illustrated in Fig.~\ref{fig:intro1}(c). 

Only a few methods are proposed with the considerations of reducing the annotations burden (e.g., semi, weakly-supervised). L2R~\cite{liu2019exploiting} facilitates the counting task by ranking the image patch. Wang \textit{et al.}~\cite{wang2019learning} introduce a synthetic crowd dataset named GCC, and the model is pre-trained on the GCC dataset and then fine-tuned on real data. One of the most relevant works for our method is ~\cite{yang2020weakly}, which proposes a soft-label network to facilitate the counting task, directly regressing the crowd number without the supervision of location labels. However, ~\cite{yang2020weakly} is a CNN-based method, which has a limited respective field. The Transformer has a global receptive field, which effectively solves the limited respective field problem of CNN-based methods once and for all. It means that the Transformer architecture is more suitable for the weakly-supervised counting task since the task aims to directly predict a total count from the whole image and rely on the global perspective. 

In this paper, we introduce two types of TransCrowd, named TransCrowd-Token and TransCrowd-GAP, respectively. TransCrowd-Token utilizes an extra learnable token to represent the count. TransCrowd-GAP adopts the global average pooling (GAP) over all items in the output sequence of Transformer-encoder to obtain the pooled visual tokens. The regression tokens or pooled visual tokens are then fed into the regression head to generate the prediction count. We empirically find that the TransCrowd-GAP can obtain more reasonable attention weight, achieve higher count accuracy, and present fast-converging compared with TransCrowd-Token. 

In summary, this work contributes to the following:
\begin{enumerate}
\item TransCrowd is the first pure transformer-based crowd counting framework. We reformulate the counting problem from a sequence-to-count perspective and propose a weakly-supervised counting method, which only utilizes the count-level annotations without the point-level information in the training phase. 

\item We provide two different types of TransCrowd, named TransCrowd-Token and TransCrowd-GAP, respectively. We observe that the TransCrowd-GAP can generate a more reasonable attention weight and reports faster converging and higher counting performance than TransCrowd-Token.

\item Extensive experiments demonstrate that the proposed method achieves state-of-the-art counting performance compared with the weakly-supervised methods. Additionally, our method has a highly competitive counting performance compared with the fully-supervised counting methods.
\end{enumerate}

\section{Related Works}

\subsection{CNN-based crowd counting.}
The CNN-based crowd counting methods can be categorized into localization-based methods and regression-based methods. The localization-based methods~\cite{ren2015faster,liu2016ssd} usually learn to predict bounding boxes for each human, relying on box-level annotations. Recently, some methods~\cite{abousamra2020localization, liu2019point, liang2021focal,xu2019autoscale,chen2021cell} try to utilize the pseudo bounding boxes based on point-level annotations or design a suitable map to realize counting and localization tasks. However, these localization-based methods usually report unsatisfactory counting performance. The mainstream of crowd counting is the density map CNN-based crowd counting methods~\cite{zhang2016single,li2018csrnet,zhang2019attentional,du2020visdrone,ma2019bayesian,jiang2020attention,xu2020dilated}, whose integral of the density map gives the total count of a crowd image. Due to the commonly heavy occlusion that existed in crowd images, multi-scale architecture is developed. Specifically, MCNN~\cite{zhang2016single} utilizes multi-size filters to extract different scale feature information. Sam \textit{et al.}~\cite{sindagi2017generating} capture the multi-scale information by the proposed contextual pyramid CNN. 
TEDNet~\cite{jiang2019crowd} assembles multiple encoding-decoding paths hierarchically to generate a high-quality density map for accurate crowd counting. Method in~\cite{ma2020learning} proposes a scale-aware probabilistic model to handle large scale variations through DPN, and each level of DPN copes with a given scale range.
Using the perspective information to diminish the scale variations is effective~\cite{shi2019revisiting,gao2019pcc,yang2020reverse}. 
PACNN~\cite{shi2019revisiting} proposes a novel generating ground truth perspective maps strategy and predicts both the perspective maps and density maps at the testing phase. Yang \textit{et al.}~\cite{yang2020reverse} propose a reverse perspective network to estimate the perspective factor of the input image and then warp the image. 
Appropriate measure matching can help to improve the counting performance. S3~\cite{lin2021direct} propose a novel measure matching based on Sinkhorn divergence, avoiding generating the density maps. UOT~\cite{ma2021learning} use unbalanced optimal transport (UOT) distance to quantify the discrepancy between two measures, outputting sharper density maps.

The Attention-based mechanism is another useful technique adopted by many methods~\cite{liu2019adcrowdnet,jiang2020attention,zhang2019attentional}. ADCrowdNet~\cite{liu2019adcrowdnet} generates an attention map for the crowd images via a network called Attention Map Generate (AMG). Jiang \textit{et al.}~\cite{jiang2020attention} propose a density attention network to generate attention masks concerning regions of different density levels. Zhang \textit{et al.}~\cite{zhang2019attentional} propose a Relation Attention Network (RANet) that utilizes local self-attention and global self-attention to capture long-range dependencies. It is noteworthy that RANet~\cite{zhang2019attentional} is actually a non-local/self-attention mechanism based on CNN instead of pure Transformers, and we utilize a pure Transformer without convolution layers.

\begin{figure*}[t]
\centering
\resizebox{0.97\textwidth}{!}{
    \includegraphics{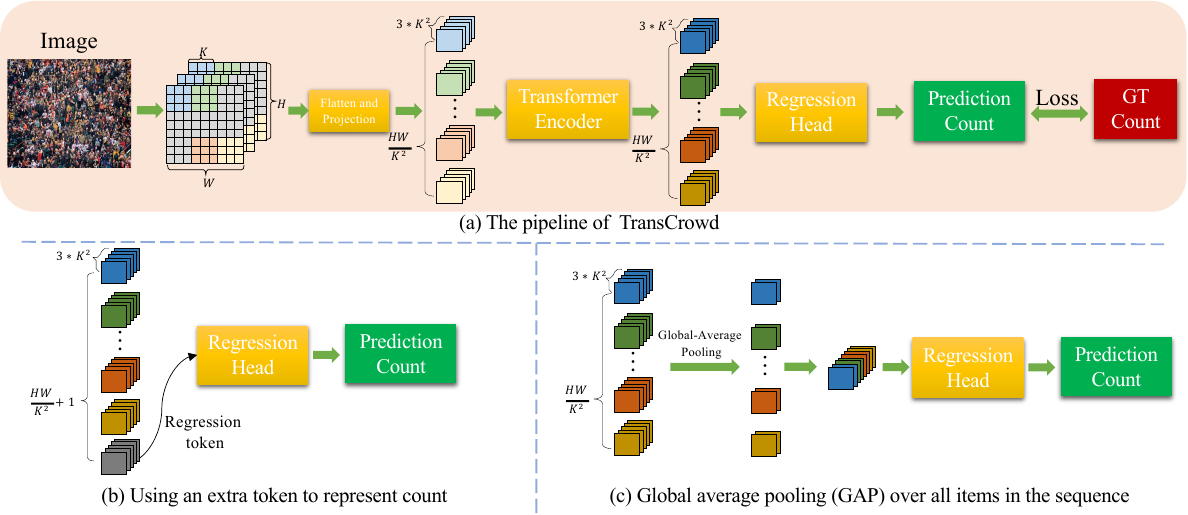}
    }
\caption{(a) The pipeline of TransCrowd. The input image is split into fixed-size patches, each of which is linearly embedded with position embeddings. Then, the feature embedding sequence is fed into the Transformer-encoder, followed by a regression head to generate the prediction count. (b) We utilize an extra token to represent the crowd count, similar to the class token in Bert~\cite{devlin2018bert} and ViT~\cite{vit}. (c) A global average pooling is adopted to pool the output visual tokens of the Transformer-encoder.}
\label{fig:pipeline}
\centering
\end{figure*}

\subsection{Weakly-supervised crowd counting.}
Only a few methods focus on counting with a lack of labeled data. 
L2R~\cite{liu2019exploiting} proposes a learning-to-rank framework based on an extra collected crowd dataset. Wang \textit{et al.}~\cite{wang2019learning} introduce a synthetic crowd scene for the pre-trained model. However, these two methods still rely on point-level annotations, which are fully-supervised instead of weakly-supervised paradigms. 

The traditional method~\cite{chan2008privacy} relies on hand-crafted features, such as GLCM and edge orientations, which are turned to be sub-optimal for this weakly-supervised counting task. MATT~\cite{lei2021towards} learns a model from a small amount of point-level annotations (fully-supervised) and a large amount of count-level annotations (weakly-supervised). 
The method in ~\cite{von2016gaussian} proposes a weakly-supervised solution based on the Gaussian process for crowd density estimation. Shang~\textit{et al.} simultaneous predicted the global count and local count. Wang \textit{et al.} directly regress the global count, and some negative samples are fed into the network to boost the robustness.
Similarly, Yang \textit{et al.}~\cite{yang2020weakly} also directly maps the images to the crowd numbers without the location supervision based on the proposed soft-label sorting network. 

However, the counting performance of these count-level weakly-supervised counting methods still does not achieve comparable results to the fully-supervised counting methods, existing massive degradation, limiting the application of weakly-supervised methods in real-world. 
Different from the previous works, the proposed TransCrowd utilizes the Transformer architecture to directly regress the crowd number, which formulates the counting problem as the sequence-to-count paradigm and achieves comparable counting performance compared with the popular fully-supervised methods.

\subsection{Visual Transformer.} 
The Transformers~\cite{vaswani2017attention}, dominating the natural language modeling~\cite{devlin2018bert, liu2019roberta}, utilize the self-attention mechanism to capture the global dependencies between input and output. Recently, many works~\cite{detr, vit, SETR, transmeettrack, pretrained} attempt to apply the Transformer into the vision task. Specifically, DETR~\cite{detr} firstly utilizes a CNN backbone to extract the visual features, followed by the Transformer blocks for the box regression and classification. 
ViT~\cite{vit} is the first which directly applies Transformer-encoder~\cite{vaswani2017attention} to sequences of images patch to realize classification task. SETR~\cite{SETR} regards semantic segmentation from a sequence-to-sequence perspective with Transformers. IPT~\cite{pretrained} develops a pre-trained model for image processing (low-level task) using the transformer architecture. 

To the best of our knowledge, we are the first to explore the pure Transformer~\cite{vaswani2017attention} to the counting task.

\section{Our Method}
The overview of our method consists of the sequence (tokens) of the image, a Transformer-encoder, and a naive regression head, as shown in Fig.~\ref{fig:pipeline}(a). Specifically, the input image is first transformed into fixed-size patches and then flatten to a sequence of vectors. The sequence is feed into the Transformer-encoder, followed by a naive regression head to generate the prediction count. 

\subsection{Image to sequence}
In general, the Transformer adopts a 1$D$ sequence of feature embeddings $Z\in \mathbb{R}^{N \times D}$ as input, where $N$ is the length of the sequence and the $D$ means the input channel size. Thus, the first step of TransCrowd is to transform the input image $I$ into a sequence of 2D flattened patches.
Specifically, given an RGB image \footnote{$H$, $W$, 3 indicate the spatial height, width and channel number, respectively.} $I \in \mathbb{R}^{H \times W \times 3}$ , we reshape the $I$ into a grid of $N$ patches, resulting in \{$x^i_p \in \mathbb{R}^{K^2 \cdot 3}| i=1,...,N$\}, where $N = \frac{H}{K} \times \frac{W}{K}$ and $K$ is the pre-defined patch size.

\subsection{Patch Embedding}
Next, we need to map the $x$ into a latent $D$-dimensional embedding feature by a learnable projection, since the transformer uses constant latent vector size  D through all of its layers, defined as

\begin{equation}
   e = [e_1;e_2\cdots; e_N] = [x^1_p E;x^2_p E;\cdots; x^N_p E], \qquad E \in \mathbb{R}^{(K^2\cdot 3) \times D},
\end{equation}
where $E$ is a learnable matrix, and $e \in \mathbb{R}^{N \times D}$ is the mapped features. Thus, we add a specific position embedding \{$p_i \in \mathbb{R}^{D} |i=1,...,N$\} into the $e$, maintaining position information, defined as:

\begin{equation}
    Z_0 = [Z^1_0;Z^2_0;\cdots; Z^N_0] = [e_1 + p_1;e_2 + p_2;\cdots; e_N + p_N],
\end{equation}
where $Z_0$ is the input of the first transformer layer.



\begin{figure}[t]
	\begin{center}
		\includegraphics[width=1\linewidth]{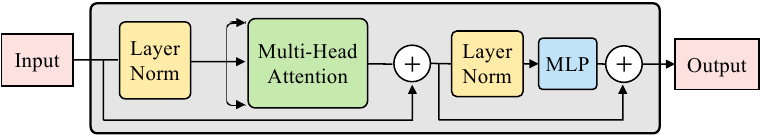}
	\end{center}
	\caption{A standard Transformer layer consists of Multi-Head Attention and MLP blocks. Meanwhile, the layer normalization (LN) and residual connections are employed.}
	\label{fig:transformer_layer}
\end{figure}

\subsection{Transformer-encoder}
We only adopt the Transformer encoder~\cite{vaswani2017attention}, without the decoder, similar to ViT~\cite{vit}. Specifically, the encoder contains $L$ layers of Multi-head self-attention ($MSA$) and Multilayer Perceptron ($MLP$) blocks. For each layer $l$, layer normalization (LN) and residual connections are employed. A stand transformer layer is shown in Fig.~\ref{fig:transformer_layer}, and the output can be written as follows:

\begin{equation}
Z'_l = MSA(LN(Z_{l-1})) + Z_{l-1},
\label{eq:encoder1}
\end{equation}

\begin{equation}
Z_l = MLP(LN(Z'_l)) + Z'_l,
\label{eq:encoder2}
\end{equation}
where $Z_l$ is the output of layer $l$. Here, the $MLP$ contains two linear layers with a GELU~\cite{hendrycks2016bridging} activation function. In particular, the first linear layer of MLP expands the feature embedding' dimension from $D$ to 4$D$, while the second layer shrinks the dimension from 4$D$ to $D$.

$MSA$ is an extension with $m$ independent self-attention ($SA$) modules: $MSA(Z_{l-1}) = [SA_1(Z_{l-1});$ $SA_2(Z_{l-1});\cdots;SA_m(Z_{l-1})]W_{O}$, where $W_{O} \in \mathbb{R}^{D \times D}$ is a re-projection matrix. At each independent $SA$, the input consists of query (Q), key (K), and value (V), which are computed from $Z^{l-1}$:
\begin{equation}
Q = Z_{l-1}W_Q, \quad K = Z_{l-1}W_K, \quad V =Z_{l-1}W_V,
\label{eq:self_attention}
\end{equation}

\begin{equation}
SA(Z_l) = softmax(\frac{QK^T}{\sqrt{D}}) V
\label{eq:self_attention}
\end{equation}
where $W_Q$/$W_K$/$W_V$ $\in \mathbb{R}^{D\times \frac{D}{m}}$ are three learnable matrices. The $softmax$ function is applied over each row of the input matrix and $\sqrt{D}$ provides appropriate normalization. 

\subsection{The input of regression head}
We introduce two different inputs for the regression heads to evaluate the effectiveness of TransCrowd. The goal of the regression head is to generate the prediction count instead of the density map. We briefly describe the two types of input.

\textbf{(1) Regression Token.} Similar to the class token in Bert~\cite{devlin2018bert} and ViT~\cite{vit}, we prepend a learnable embedding named regression token to the input sequence $Z_0$, as shown in Fig.~\ref{fig:pipeline}(b). This architecture forces the self-attention to spread information between the patch tokens and the regression token, making the regression token contain overall semantic crowd information. The regression head is implemented by MLP containing two linear layers. We refer to the TransCrowd with the extra regression token as TransCrowd-Token. 

\textbf{(2) Global average pooling.} We apply the global average pooling (GAP) to shrink the sequence length, as shown in Fig.~\ref{fig:pipeline}(c). Similar to TransCrowd-Token, two linear layers are used for the regression head. We refer to the TransCrowd with global average pooling as TransCrowd-GAP. The global average pooling can effectively maintain the useful semantic crowd information in patch tokens. We find that using pooled visual tokens will generate richer discriminative semantic crowd patterns and achieve better counting performance than using the extra regression token, the detailed discussion listed in Sec.~\ref{sec:visual}.

We utilize $L_1$ loss to measure the difference between prediction and ground truth:
\begin{equation}
L_1=\frac{1}{M}\sum_{i=1}^{M}\left |P_{i}-G_{i} \right|,
\label{eq:self_attention}
\end{equation}
where $P_{i}$ and $G_{i}$ are the prediction crowd number and the corresponding ground truth of $i$-th image, respectively. $M$ is the batch size of training images.

\section{Experiments}
\subsection{Implementation details}

The Transformer-encoder is similar to ViT~\cite{vit}, which contains 12 transformer layers, and each $MSA$ consists of 12 $SA$. We utilize the fixed $H$ and $W$, 
both of which are set as 384.
We set $K$ as 16, which means $N$ is equal to 576. 
We use Adam~\cite{kingma2014adam} to optimize our model, in which the learning rate and weight decay are set to 1e-5 and 1e-4, respectively.
The weights pre-trained on ImageNet are used to initialize the Transformer-encoder. 
During training, the widely adopted data augmentation strategies are utilized, including random horizontal flipping and grayscaling. 
Due to some datasets having various resolution images, we resize all the images into the size of 1152 $\times$ 768. Each resized image can be regarded as six independent sub-image, and the resolution of each sub-image is 384 $\times$ 384. We set the batch size as 24 and use a V100 GPU for the experiments. Code and models will be released for the community to follow.

\subsection{Dataset}
\textbf{NWPU-Crowd}~\cite{gao2020nwpu}, a large-scale and challenging dataset, consists of 5,109 images, 2,133,375 instances annotated elaborately. To be specific, the images are randomly split into three parts, including training, validation, and testing sets, which contain 3,109, 500, and 1,500 images, respectively.

\textbf{JHU-CROWD++}~\cite{sindagi2020jhu} contains 2,722 training images, 500 validation images, and 1,600 testing images, collected from diverse scenarios. The total number of people in each image ranges from 0 to 25,791.

\textbf{UCF-QNRF}~\cite{idrees2018composition} contains 1,535 images captured from unconstrained crowd scenes with about one million annotations. It has a count range of 49 to 12,865, with an average count of 815.4. Specifically, the training set consists of 1,201 images, and the testing set consists of 334 images.

\begin{table*}[t]
\normalfont
\centering
\setlength{\tabcolsep}{3mm}
\resizebox{0.97\textwidth}{!}{
\begin{tabular}{ |l|c|cc|cc|cc|cc|cc| }
 \hline
{\multirow{2}{*}{Method}} &{\multirow{2}{*}{Year}}&\multicolumn{2}{c|}{Training label}&\multicolumn{2}{c|}{UCF-QNRF}  & \multicolumn{2}{c|}{Part A} & \multicolumn{2}{c|}{Part B} \\
\cline{3-10}
&&Location&Crowd number& MAE & MSE & MAE & MSE&MAE&MSE \\
 \hline
    MCNN~\cite{zhang2016single} &CVPR 16&$\surd$&$\surd$&277.0&426.0& 110.2 & 173.2 & 26.4 & 41.3  \\
    CL~\cite{idrees2018composition}&ECCV 18&$\surd$&$\surd$&132.0&191.0&-&-&-&-\\
    CSRNet~\cite{li2018csrnet} &CVPR 18&$\surd$&$\surd$&-&-& 68.2 & 115.0  & 10.6&16.0 \\
    L2R~\cite{liu2019exploiting} &TPAMI 19&$\surd$&$\surd$&124.0&196.0& 73.6 & 112.0 & 13.7 & 21.4  \\
    CFF~\cite{shi2019counting} &ICCV 19 &$\surd$&$\surd$&- &-&65.2 &109.4 &	7.2 & 12.2 \\
    PGCNet~\cite{yan2019perspective} &ICCV19&$\surd$&$\surd$&-&-&57.0&\textbf{86.0}&8.8&13.7\\
    TEDnet~\cite{jiang2019crowd}&CVPR 19&$\surd$&$\surd$&113.0&188.0&64.2&109.1&8.2&12.8\\
    BL~\cite{ma2019bayesian} &ICCV 19 &$\surd$&$\surd$&88.7&154.8& 62.8& 101.8& 7.7 & 12.7\\
    ASNet~\cite{jiang2020attention}&CVPR 20&$\surd$&$\surd$&91.5&159.7&57.7&90.1&-&-\\
    LibraNet~\cite{liu2020weighing} &ECCV20&$\surd$&$\surd$&88.1&143.7&\textbf{55.9}&97.1&7.3&11.3\\
    NoisyCC~\cite{wan2020modeling}&NeurIPS 20&$\surd$&$\surd$&85.8&150.6&61.9&99.6&7.4&11.3\\
    DM-Count~\cite{wan2020modeling}&NeurIPS 20&$\surd$&$\surd$&85.6&148.3&59.7&95.7&7.4&11.8\\
    Method in~\cite{ma2020learning}&MM 20&$\surd$&$\surd$&84.7&147.2&58.1&91.7&6.5&\textbf{10.1}\\
    S3~\cite{lin2021direct}&IJCAI 21&$\surd$&$\surd$&\textbf{80.6}&\textbf{139.8}&57.0&96.0&\textbf{6.3}&10.6\\
    UOT~\cite{ma2021learning}&AAAI 21&$\surd$&$\surd$&83.3&142.3&58.1&95.9&6.5&10.2\\
    \hline
    Method in~\cite{yang2020weakly}*&ECCV 20&$-$&$\surd$ &-&-& 104.6 & 145.2  & 12.3  &21.2\\
    MATT~\cite{lei2021towards}*&PR 21&$-$&$\surd$ &-&-& 80.1 & 129.4 & 11.7 &17.5\\
     \textbf{TransCrowd-Token (ours)*}&-&$-$&$\surd$&98.9&176.1&69.0&116.5&10.6&19.7 \\
     \textbf{TransCrowd-GAP (ours)*}&-&$-$&$\surd$&\textbf{97.2}&\textbf{168.5}&\textbf{66.1}&\textbf{105.1}&\textbf{9.3}&\textbf{16.1} \\
    \hline
\end{tabular}}
\caption{Quantitative comparison (in terms of MAE and MSE) of the proposed method and some popular methods on three widely adopted benchmark datasets. * represents the weakly-supervised method.}
\label{tab:qab_maemse}
\end{table*}

\begin{table*}[t]
\normalfont
\centering
\setlength{\tabcolsep}{2.2mm}
\resizebox{0.97\textwidth}{!}{
\begin{tabular}{|l|c|cc|cc|cc|cc|cc|cc|cc|cc|cc|cc|}
	\hline
	\multirow{3}{*}{Method}&\multirow{3}{*}{Year}&\multicolumn{2}{c|}{\multirow{2}{*}{Training label}}  &\multicolumn{8}{c|}{Val set}\\
	 \cline{5-12} 
	&&&& \multicolumn{2}{c|}{Low}  & \multicolumn{2}{c|}{Medium}& \multicolumn{2}{c|}{High}&\multicolumn{2}{c|}{Overall}\\
	 \cline{3-12} 
	&&Location&Crowd number & MAE & MSE & MAE & MSE& MAE& MSE& MAE & MSE\\ \hline
	MCNN~\cite{zhang2016single} &CVPR16&$\surd$&$\surd$& 90.6& 202.9& 125.3& 259.5& 494.9 & 856.0 & 160.6& 377.7 \\  
	CMTL~\cite{sindagi2017cnn} &AVSS17&$\surd$&$\surd$& 50.2& 129.2& 88.1 & 170.7& 583.1 & 986.5& 138.1& 379.5\\  
	DSSI-Net~\cite{liu2019crowd}&ICCV19&$\surd$&$\surd$& 50.3& 85.9 & 82.4 & 164.5& 436.6 & 814.0 & 116.6& 317.4 \\  
	CAN~\cite{liu2019context}&CVPR19&$\surd$&$\surd$& 34.2& 69.5 & 65.6 & 115.3& 336.4 & \textbf{619.7} & 89.5 & 239.3\\  
	SANet~\cite{cao2018scale} &ECCV18&$\surd$&$\surd$& 13.6& 26.8 & 50.4 & 78.0 & 397.8 & 749.2& 82.1 & 272.6 \\ 
	CSRNet~\cite{li2018csrnet} &CVPR18&$\surd$&$\surd$& 22.2& 40.0 & 49.0 & 99.5 & 302.5 & 669.5 & 72.2 & 249.9 \\  
	CG-DRCN~\cite{sindagi2020jhu}&PAMI20&$\surd$&$\surd$& 17.1& 44.7 & 40.8 & \textbf{71.2}      & 317.4 & 719.8 & 67.9 & 262.1 \\
	MBTTBF~\cite{sindagi2019multi} &ICCV19&$\surd$&$\surd$& 23.3& 48.5 & 53.2 & 119.9& 294.5 & 674.5 & 73.8 & 256.8\\  
	SFCN~\cite{wang2019learning} &CVPR19&$\surd$&$\surd$& 11.8& 19.8 & \textbf{39.3}& 73.4 & 297.3 & 679.4&62.9 & 247.5 \\  
	BL~\cite{ma2019bayesian}& ICCV19&$\surd$&$\surd$& \textbf{6.9}&\textbf{10.3}& 39.7 & 85.2 &\textbf{279.8}& 620.4& \textbf{59.3} & \textbf{229.2}\\  
	\hline
	\textbf{TransCrowd-Token (ours)*}&-&-&$\surd$&7.1&10.7&\textbf{33.3}&\textbf{54.6}&302.5&557.4&58.4&201.1\\
	\textbf{TransCrowd-GAP (ours)*} &-&-&$\surd$&\textbf{6.7}&\textbf{9.5}&34.5&55.8&\textbf{285.9}&\textbf{532.8}&\textbf{56.8}&\textbf{193.6}\\
	\hline
    \end{tabular}}
\caption{Quantitative results on the JHU-Crowd++ (val set) dataset. "Low", "Medium" and "High" respectively indicates three categories based on different ranges:[0,50], (50,500], and \textgreater 500. * represents the weakly-supervised crowd counting methods.}
\label{tab:jhu_val_maemse}
\end{table*}

\textbf{ShanghaiTech}~\cite{zhang2016single} contains 1,198 crowd images with 330,165 annotations. The images of the dataset are divided into two parts: Part A and Part B. In particular, Part A contains 300 training images and 182 testing images, and Part B consists of 400 training images and 316 testing images. 

\textbf{UCF\_CC\_50}~\cite{idrees2013multi} is a small dataset for dense crowd counting, which just contains 50 images with an average of 1,280 individuals per image. The images are captured in a diverse set of events, and the pedestrians count of each image range between 94 and 4,543.

\textbf{WorldExpo'10}~\cite{zhang2015cross} contains 1,132 surveillance videos from 108 cameras. The training set consists of 3,380 images captured from 103 different scenes, and the testing set contains 600 images from 5 scenes. There are 199,923 annotations labelled in the whole 3,980 images.

\subsection{Evaluation Metrics}
We choose Mean Absolute Error (MAE) and Mean Squared Error (MSE) to evaluate the counting performance:

\begin{equation}
MAE=\frac{1}{N}\sum_{i=1}^{N}\left |P_{i}-G_{i} \right|,
MSE=\sqrt{\frac{1}{N}\sum_{i=1}^{N}\left |P_{i}-G_{i} \right|^{2}},
\label{eq:mse}
\end{equation}
where $N$ is the number of testing images, $P_{i}$ and $G_{i}$ are the predicted and ground truth count of the $i$-th image, respectively.

\begin{table*}[t]
\normalfont
\centering
\setlength{\tabcolsep}{2mm}
\resizebox{0.97\textwidth}{!}{
\begin{tabular}{|l|c|cc|cc|cc|cc|cc|cc|cc|cc|cc|}
	\hline
	\multirow{3}{*}{Method}&\multirow{3}{*}{Year}&\multicolumn{2}{c|}{\multirow{2}{*}{Training label}}  &\multicolumn{8}{c|}{Testing set}\\
	 \cline{5-12} 
	&&&& \multicolumn{2}{c|}{Low}  & \multicolumn{2}{c|}{Medium}& \multicolumn{2}{c|}{High}&\multicolumn{2}{c|}{Overall}\\
	 \cline{3-12} 
	&&Location&Crowd number& MAE & MSE & MAE & MSE& MAE& MSE& MAE & MSE \\ \hline
	MCNN~\cite{zhang2016single} &CVPR16&$\surd$&$\surd$& 97.1 & 192.3& 121.4& 191.3& 618.6& 1,166.7 & 188.9& 483.4 \\  
	CMTL~\cite{sindagi2017cnn} &AVSS17&$\surd$&$\surd$& 58.5& 136.4& 81.7 & 144.7 & 635.3& 1,225.3 & 157.8 &490.4\\  
	DSSI-Net~\cite{liu2019crowd}&ICCV19&$\surd$&$\surd$& 53.6& 112.8  & 70.3& 108.6  & 525.5& 1,047.4 & 133.5  & 416.5\\  
	CAN~\cite{liu2019context}&CVPR19&$\surd$&$\surd$& 37.6& 78.8& 56.4& 86.2& 384.2& 789.0& 100.1& 314.0\\  
	SANet~\cite{cao2018scale} &ECCV18&$\surd$&$\surd$& 17.3& 37.9& 46.8& 69.1& 397.9& 817.7& 91.1& 320.4\\ 
	CSRNet~\cite{li2018csrnet} &CVPR18&$\surd$&$\surd$& 27.1& 64.9& 43.9& 71.2& 356.2& 784.4& 85.9& 309.2\\  
	CG-DRCN~\cite{sindagi2020jhu}&PAMI20&$\surd$&$\surd$& 19.5& 58.7& 38.4& 62.7& 367.3& 837.5&82.3& 328.0\\
	MBTTBF~\cite{sindagi2019multi} &ICCV19 &$\surd$&$\surd$& 19.2& 58.8& 41.6& 66.0& 352.2& 760.4& 81.8&299.1\\  
	SFCN~\cite{wang2019learning} &CVPR19&$\surd$&$\surd$& 16.5& 55.7& 38.1& 59.8& 341.8&758.8 & 77.5&297.6\\  
	BL~\cite{ma2019bayesian}& ICCV19&$\surd$&$\surd$&\textbf{10.1}&32.7& 34.2&54.5& 352.0& 768.7&75.0& 299.9\\  
	UOT~\cite{ma2021learning}& AAAI21&$\surd$&$\surd$&11.2&\textbf{26.2}&\textbf{28.7}&\textbf{45.3}& \textbf{274.1}& \textbf{648.2}&60.5& 252.7\\  
	S3~\cite{lin2021direct}& IJCAI21&$\surd$&$\surd$&-&-&-&-&-& -&\textbf{59.4}& \textbf{244.0}\\  
	\hline
	\textbf{TransCrowd-Token (ours)*}&-&-&$\surd$&8.5&23.2&\textbf{33.3}&\textbf{71.5}&368.3&816.4&76.4&319.8\\
	\textbf{TransCrowd-GAP (ours)*} &-&-&$\surd$&\textbf{7.6}&\textbf{16.7}&34.8&73.6&\textbf{354.8}&\textbf{752.8}&\textbf{74.9}&\textbf{295.6} \\
	\hline
    \end{tabular}}
\caption{Quantitative results on the JHU-Crowd++ (testing set) dataset. "Low", "Medium" and "High" respectively indicates three categories based on different ranges:[0,50], (50,500], and \textgreater 500. * represents the weakly-supervised crowd counting methods.}
\label{tab:jhu_test_maemse}
\end{table*}

\begin{table*}[t]
\footnotesize
\centering
\setlength{\tabcolsep}{1mm}
\resizebox{0.97\textwidth}{!}{
\begin{tabular}{|c|c|cc|c|c|c|c|c|c|c|}
	\hline
	\multirow{3}{*}{Method}&\multirow{3}{*}{Year}&\multicolumn{2}{c|}{\multirow{2}*{Training label}} &\multicolumn{2}{c|}{Val set}&\multicolumn{4}{c|}{Testing set}     \\
	\cline{5-10}
	&&&&\multicolumn{2}{c|}{Overall}&\multicolumn{2}{c|}{Overall} &\multicolumn{2}{c|}{Scene Level (only MAE)} \\
	\cline{3-10}
	&&Location&Crowd number& MAE &MSE  & MAE &MSE  &Avg. & $S0 \sim S4$   \\

	\hline
	C3F-VGG~\cite{gao2019c}  &Tech19&$\surd$&$\surd$&105.79 &504.39& 127.0 & 439.6  & {666.9} & 140.9/26.5/58.0/307.1/2801.8    \\
	CSRNet~\cite{li2018csrnet}  &CVPR18&$\surd$&$\surd$&104.89 &433.48&121.3 & 387.8 & 522.7 & 176.0/35.8/59.8/285.8/2055.8    \\
	PCC-Net-VGG ~\cite{gao2019pcc} &CVPR19&$\surd$&$\surd$&100.77 &573.19& 112.3 & 457.0  & 777.6 & 103.9/13.7/42.0/259.5/3469.1   \\			
	CAN~\cite{liu2019context}  &CVPR19&$\surd$&$\surd$&93.58 &489.90& {106.3} & 386.5  &612.2 & 82.6/14.7/46.6/269.7/2647.0  \\
	SFCN\dag~\cite{wang2019learning}  &CVPR19&$\surd$&$\surd$&95.46 &608.32& 105.7 & 424.1  & 712.7 & 54.2/14.8/44.4/249.6/3200.5   \\
	BL~\cite{ma2019bayesian}  &ICCV19&$\surd$&$\surd$&93.64 &470.38&105.4  &454.2   & 750.5 & 66.5/8.7/41.2/249.9/3386.4\\
    KDMG~\cite{wan2020kernel} &PAMI20&$\surd$&$\surd$&-&-&100.5&415.5&632.7&77.3/10.3/38.5/259.4/2777.9\\
	NoisyCC~\cite{wan2020modeling}  & NeurIPS20&$\surd$&$\surd$ &-&-&96.9&534.2&608.1&218.7/10.7/35.2/203.2/2572.8 \\
	DM-Count ~\cite{wang2020distribution} & NeurIPS20&$\surd$&$\surd$ &\textbf{70.5}&\textbf{357.6}&88.4&388.6&\textbf{498.0}&146.6/7.6/31.2/228.7/2075.8 \\
	S3~\cite{lin2021direct} & IJCAI21&$\surd$&$\surd$ &-&-&87.8&387.5&566.5&80.7 / 7.9 / 36.3 / 212.0 / 2495.4 \\
	UOT~\cite{ma2021learning} & AAAI21&$\surd$&$\surd$ &-&-&\textbf{83.5}&\textbf{346.9}&-&- \\
\hline
\textbf{TransCrowd-Token (ours)*}  &-&$-$&$\surd$& \textbf{88.2}&446.9&119.6&463.9 & \textbf{736.0}&88.0/12.7/47.2/311.2/3216.1  \\
\textbf{TransCrowd-GAP (ours)*}  &-&$-$&$\surd$&88.4&\textbf{400.5}& \textbf{117.7} & \textbf{451.0} & 737.8 & 69.3/12.8/46.0/309.0/3252.2 \\
\hline
\end{tabular}}
\caption{Comparison of the counting performance on the NWPU-Crowd. $S0\!\sim\!S4$ respectively indicate five categories according to the different number range: $0$, $(0, 100]$, $(100, 500]$, $(500, 5000]$, $\textgreater5000$. * represents the weakly-supervised crowd counting methods.}
\label{tab:NWPU_maemse}
\end{table*} 

\begin{table*}[t]
\footnotesize
\centering
\setlength{\tabcolsep}{1mm}
\begin{tabular}{ |l|c|c|c|c| }
\hline
{\multirow{2}{*}{Method}} &\multicolumn{2}{c|}{Training label} &{\multirow{2}{*}{MAE}}&{\multirow{2}{*}{MSE}}\\
\cline{2-3}
&Location&Crowd number&&\\
\hline
MCNN~\cite{zhang2017fcn}&$\surd$&$\surd$&377.6&509.1\\
CSRNet~\cite{li2018csrnet}&$\surd$&$\surd$&266.1&397.5\\
ADCrowdNet~\cite{liu2019adcrowdnet}&$\surd$&$\surd$&\textbf{257.1}&\textbf{363.5}\\
\hline
MATT~\cite{lei2021towards}*&$-$&$\surd$&355.0 &550.2\\
\textbf{TransCrowd-Token (ours)*} &$-$&$\surd$& 288.9&407.6\\
\textbf{TransCrowd-GAP (ours)*} &$-$&$\surd$& \textbf{272.2}&\textbf{395.3}\\
\hline
\end{tabular}
\caption{The performance comparison on the UCF\_CC\_50 dataset. * represents the weakly-supervised crowd counting methods.}
\label{tab:ucf50}
\end{table*}

\begin{table*}[t]
\footnotesize
\centering
\setlength{\tabcolsep}{1mm}
	\begin{tabular}{|l|c|c|c|c|c|c|c|c|}
		\hline
		\multirow{2}[4]{*}[6pt]{\textbf{Method}} &\multicolumn{2}{c|}{Training label}& \multicolumn{6}{c|}{\textbf{MAE}} \\
		\cline{2-9}
		&Location&Crowd number& \textbf{S1} & \textbf{S2} & \textbf{S3} & \textbf{S4} & \textbf{S5} & \textbf{Ave} \\
		 \hline
		M-CNN \cite{zhang2016single} &$\surd$&$\surd$& 3.4   & 20.6  & 12.9  & 13.0    & 8.1   & 11.6 \\ 
		CP-CNN\cite{sindagi2017generating}   &$\surd$&$\surd$& 2.9 & 14.7 & 10.5 & 10.4 & 5.8  & 8.8 \\
        Liu et al.\cite{liu2018decidenet}&$\surd$&$\surd$& \textbf{2.0} & 13.1 & 8.9 & 17.4 &  4.8 & 9.2 \\
		IC-CNN\cite{ranjan2018iterative} &$\surd$&$\surd$   & 17.0 & 12.3 & 9.2 & \textbf{8.1} & 4.7   & 10.3 \\
		CSR-Net\cite{li2018csrnet}  &$\surd$&$\surd$  & 2.9 & 11.5 & \textbf{8.6} & 16.6 & 3.4  & 8.6  \\
		SA-Net\cite{cao2018scale}   &$\surd$&$\surd$  & 2.6 & 13.2 & 9.0 & 13.3 & \textbf{3.0}  & 8.2  \\
		LSC-CNN\cite{sam2020locate}  &$\surd$&$\surd$  & 2.9 & \textbf{11.3} & 9.4 & 12.3 & 4.3  & \textbf{8.0}  \\
		\hline
		MATT\cite{lei2021towards}*  &$-$&$\surd$  & 3.8 & \textbf{13.1} & 10.4 & 15.9 & 5.3  & 9.7  \\
		\textbf{TransCrowd-Token\textbf{ (ours)*}}  &$-$&$\surd$ &2.3 & 14.2 & 9.9 & 14.0 & \textbf{4.3} & 8.9    \\
		\textbf{TransCrowd-GAP\textbf{ (ours)*}}  &$-$&$\surd$  &\textbf{2.1}  & 13.3 & \textbf{8.9} & \textbf{13.8} & 4.4& \textbf{8.5}   \\
		\hline
	\end{tabular}
	\caption{Comparison results of different methods on 5 scenes in the WorldExpo'10 dataset. * represents the weakly-supervised crowd counting methods.}
	\label{tab:worldexpo}%
\end{table*}

\begin{table}[t]
\small
\footnotesize
\centering
\resizebox{0.47\textwidth}{!}{
\begin{tabular}{ |l|c|c|c|c| }
\hline
{\multirow{2}{*}{Method}} &\multicolumn{2}{c|}{Training label} &{\multirow{2}{*}{MAE}}&{\multirow{2}{*}{MSE}}\\
\cline{2-3}
&Location&Crowd number&&\\
\hline
FCN-HA~\cite{zhang2017fcn}&$\surd$&$\surd$&4.21&-\\
CSRNet~\cite{li2018csrnet}&$\surd$&$\surd$&3.56&-\\
ADCrowdNet~\cite{liu2019adcrowdnet}&$\surd$&$\surd$&\textbf{2.44}&-\\
\hline
\textbf{TransCrowd-Token (ours)*} &$-$&$\surd$& 3.28&4.80\\
\textbf{TransCrowd-GAP (ours)*} &$-$&$\surd$& \textbf{3.23}&\textbf{4.66}\\
\hline
\end{tabular}}
\caption{The performance comparison on the Trancos dataset.  * represents the weakly-supervised crowd counting methods.}
\label{tab:trancos}
\end{table}

\begin{figure*}[t]
\centering
\resizebox{0.97\textwidth}{!}{
    \includegraphics{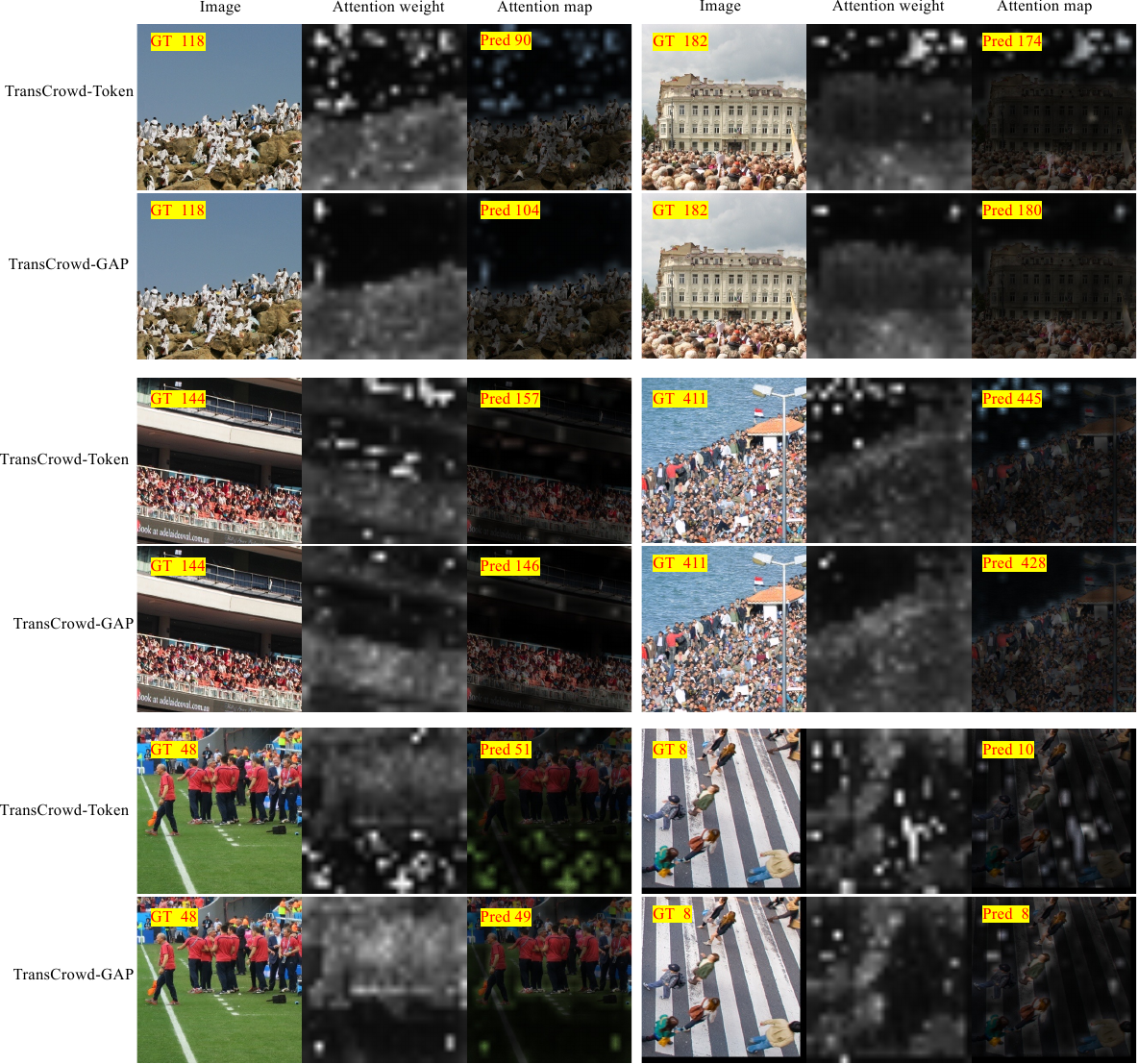}
    }
\caption{Examples of attention maps from TransCrowd-Token and TransCrowd-GAP. TransCrowd-GAP generates more reasonable attention weights compared with TransCrowd-Token. }
\label{fig:attention_map}
\centering
\end{figure*}

\section{Results}
We conduct extensive experiments to demonstrate the effectiveness of the proposed weakly-supervised crowd counting method on five popular benchmarks. For each dataset, we divide the existing methods into fully-supervised methods (based on point-level annotations) and weakly-supervised methods (based on count-level annotations). 

\textbf{Compared with the weakly-supervised counting methods.} 
Our method achieves state-of-the-art counting performance on all the conducted datasets, as listed in Tab.~\ref{tab:qab_maemse} - Tab.~\ref{tab:trancos}. Specifically, on ShanghaiTech part A, our TransCrowd-GAP improves 17.5\% in MAE and 18.8\% in MSE compared with MATT~\cite{lei2021towards}, improves 36.8\% in MAE and 27.6\% in MSE compared with~\cite{yang2020weakly}. On ShanghaiTech part B, TransCrod-GAP improves 20.5\% in MAE and 8.0\% in MSE compared with MATT~\cite{lei2021towards}, improves 24.4\% in MAE and 24.1\% in MSE compared with ~\cite{yang2020weakly}. Besides, the proposed TransCrowd-Token also achieves significant improvement compared with MATT~\cite{lei2021towards} and~\cite{yang2020weakly} in terms of MAE and MSE, and only the proposed methods report counting performance close to the fully-supervised methods.
Note that MATT~\cite{lei2021towards} still applies a small number of images, which contain point-level annotations for training.
 
\textbf{Compared with the fully-supervised counting methods.} Although it is unfair to compare the fully-supervised and weakly-supervised crowd counting methods, our method still achieves highly competitive performance on the five counting datasets, as shown in Tab.~\ref{tab:qab_maemse} - Tab.~\ref{tab:trancos}. An impressive phenomenon is that the proposed method even surpasses some popular fully-supervised methods. For example, as shown in Tab.~\ref{tab:jhu_test_maemse},  our TransCrowd-GAP brings 11.0 MAE and 13.6 MSE improvement compared with CSRNet~\cite{li2018csrnet} on the JHU-Crowd++ (testing set) dataset. BL~\cite{ma2019bayesian}, a recent strong counting method, achieves 75.0 in MAE and 299.9 in MSE, one of the state-of-the-art methods on the JHU-Crowd++ (testing set) dataset, while our TransCrowd-GAP improves 0.1 MAE and 4.3 MSE, respectively. Besides, from the results on UCF-QNRF, ShanghaiTech, and NWPU-Crowd datasets, we can also observe that our method achieve significant improvement compared to some popular fully-supervised methods (e.g., MCNN~\cite{zhang2016single}, CSRNet~\cite{li2018csrnet}, L2R~\cite{liu2019exploiting}). We think the reasons why the proposed method outperforms some fully supervised methods in the NWPU-Crowd and JHU-Crowd++ may be two-fold. First, the Transformer is beneficial to capture the long-range dependence, and these two datasets contain many large-scale persons. The proposed TransCrowd can effectively learn the global crowd semantic feature representation. However, some state-of-the-art methods (e.g., BL~\cite{ma2019bayesian}) utilize a fixed Gaussian kernel for these datasets, and the fixed Gaussian kernel can not effectively cover the large scale variations. Second, Dosovitskiy~\textit{et al.}~\cite{vit} prove that the CNNs outperform Transformer on small datasets (despite regularization optimization), but with the larger datasets, Transformer overtakes. For instance, the NWPU-Crowd is a large dataset containing 5,190 images, which may help the Transformer better fit the dataset.
These impressive results further demonstrate the effectiveness of the proposed method and indicate point-level annotations are not entirely necessary for the counting task. 

\begin{table}[tp]
\small
\footnotesize
\centering
\resizebox{0.47\textwidth}{!}{
\begin{tabular}{ |l|c|c|c|c|c|c| }
\hline
Method & Resolution &Parameters &Backbone &FPS \\
 \hline
CSRNET~\cite{li2018csrnet} & 384 $\times$ 384 &16.2 M&VGG16& 21.67 \\
BL~\cite{ma2019bayesian}& 384 $\times$ 384 &21.6 M&VGG19 &45.66 \\
\hline
\textbf{TransCrowd-Token} & 384 $\times$ 384 & 86.8 M &Transformer & 46.41 \\
\textbf{TransCrowd-GAP} & 384 $\times$ 384 & 90.4 M&Transformer & 46.73 \\
 \hline
\end{tabular}}
\caption{Comparison with BL~\cite{ma2019bayesian} and CSRNET~\cite{li2018csrnet} using the same input image resolution on a Titan XP.}
\label{tab:runtime}
\end{table}

\begin{table*}[tp]

	\centering
	\setlength{\tabcolsep}{2.5mm}
	\resizebox{0.96\textwidth}{!}{
	\begin{tabular}{ |l|c|cc|cc|cc|cc| }
        \hline
        {\multirow{2}{*}{Method}}&{\multirow{2}{*}{Year}}&\multicolumn{2}{c|}{Training label}&\multicolumn{2}{c|}{\multirow{1}{*}{None}}&\multicolumn{2}{c|}{\multirow{1}{*}{Pre-ImgNet}}&\multicolumn{2}{c|}{\multirow{1}{*}{Pre-GCC}}\\
        \cline{3-4}\cline{5-10}
         &&Location & Crowd number&MAE&MSE&MAE&MSE&MAE&MSE\\
        \hline
        CSRNet~\cite{li2018csrnet} &CVPR18&$\surd$&$\surd$&\textbf{120.0}&\textbf{179.4}&68.2&115.0&67.4&112.3\\
        \textbf{TransCrowd-Token (ours)*} &-&$-$&$\surd$&142.0&212.5&69.0&116.5&67.2&111.9 \\
        \textbf{TransCrowd-GAP (ours)*} &-&$-$&$\surd$&139.9&231.0&\textbf{66.1}&\textbf{105.1}&\textbf{63.8}&\textbf{102.3} \\
        \hline
        \end{tabular}}
	\caption{The fine-tuning CSRNet's and TransCrowd-GAP's results on ShanghaiTech part A dataset by using three different pre-trained strategies. }
	\label{tab:pretrained}%
\end{table*}%

\begin{figure}[t]
	\begin{center}
		\includegraphics[width=1.0\linewidth]{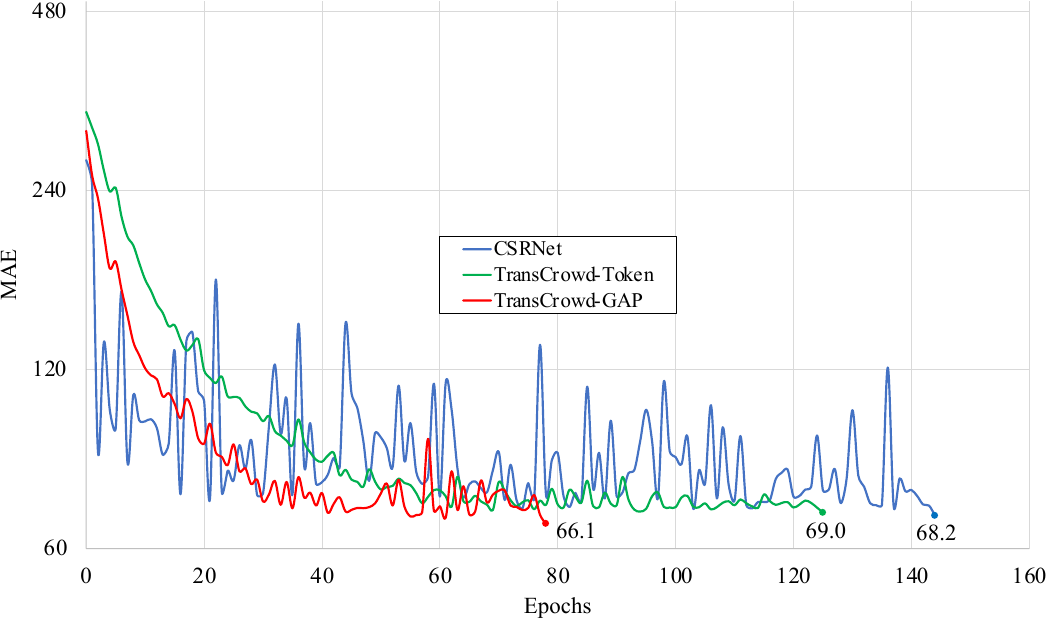}
	\end{center}
	\caption{Convergence curves of CSRNet, TransCrowd-Token, and TransCrowd-GAP on ShanghaiTech part A dataset. The proposed TransCrowd-GAP achieves the best counting performance and is fast-converging.}
	\label{fig:convergence_curves}
\end{figure}

\begin{table*}[tp]
\centering
\setlength{\tabcolsep}{1.5mm}
\resizebox{0.97\textwidth}{!}{
\begin{tabular}{ |l|c|cc|cc|cc|cc|cc|cc| }
    \hline
    {\multirow{2}{*}{Method}}&{\multirow{2}{*}{Year}}&\multicolumn{2}{c|}{Training label}&\multicolumn{2}{c|}{Part B$\rightarrow$Part A}&\multicolumn{2}{c|}{Part A$\rightarrow$Part B}&\multicolumn{2}{c|}{QNRF$\rightarrow$Part A}&\multicolumn{2}{c|}{QNRF$\rightarrow$Part B}\\
    \cline{3-12}
     &&Location & Crowd number& MAE & MSE&MAE&MSE&MAE&MSE&MAE&MSE\\
    \hline
    MCNN~\cite{zhang2016single} &CVPR16&$\surd$&$\surd$& 221.4&357.8 & 85.2&142.3&- &-&- &- \\
    D-ConvNet~\cite{shi2018crowd}&ECCV18&$\surd$&$\surd$& \textbf{140.4}&\textbf{226.1}&49.1&99.2& -&-&- &- \\
    RRSP\cite{wan2019residual} &CVPR19&$\surd$&$\surd$& -&-&\textbf{40.0}&\textbf{68.5}& -&-&- &-\\
    BL~\cite{ma2019bayesian} &ICCV19&$\surd$&$\surd$&-&-&-&-&\textbf{69.8}&\textbf{123.8}&\textbf{15.3}&\textbf{26.5}\\
    \hline
    \textbf{TransCrowd-GAP (ours)*} &-&$-$&$\surd$&\textbf{141.3}&\textbf{258.9}&\textbf{18.9}& \textbf{31.1}&\textbf{78.7} &\textbf{122.5} &\textbf{13.5}&\textbf{21.9}\\
    \hline
    \end{tabular}}
	\caption{Experimental results on the transferability of different methods under cross-dataset evaluation. }
	\label{tab:transfer}%
\end{table*}%

\section{Analysis}
\label{sec:visual}

\subsection{The input of regression head}
We introduce two different inputs for the regression head. Specifically, TransCrowd-Token utilizes an extra learnable regression token to perform counting, similar to the class token in Bert~\cite{devlin2018bert} and ViT~\cite{vit}. TransCrowd-GAP utilizes global average pooling to obtain the pooled visual tokens for count prediction. The result of TransCrowd-Token and TransCrowd-GAP are listed in Tab.~\ref{tab:qab_maemse} - Tab.~\ref{tab:trancos}. We find that the results of TransCrowd-GAP are better TransCrowd-Token in all conducted datasets. For example, TransCrowd-GAP outperforms TransCrowd-Token by 2.8 MAE and 11.4 MSE on the ShanghaiTech Part A dataset, a significant improvement. TransCrowd-GAP also has steady improvement on ShanghaiTech part B, a sparse crowd dataset. 
Based on the superior performance, we hope the researchers can design a more reasonable regression head based on the Transformer-encoder in the future.

\subsection{Visualizations}

To further investigate the proposed TransCrowd, we provide qualitative comparison results in Fig.~\ref{fig:attention_map} to understand what the Transformer attends to.
We observe that both TransCrowd-Token and TransCrowd-GAP can successfully focus on the crowd region, which demonstrates the effectiveness of both methods.
Moreover, the TransCrowd-GAP generates a more reasonable attention map compared with the TransCrowd-Token. Specifically, the TransCrowd-Token may pay more attention to the background, leading to amplifying the counting error. This observation explains why the result of TransCrowd-GAP is better than TransCrowd-Token.

\subsection{Convergence curves}
We further compare the convergence curves between the popular fully-supervised method (CSRNet~\cite{li2018csrnet}) and the proposed TransCrowd. Detailed convergence curves are shown in Fig.~\ref{fig:convergence_curves}. 
Based on the convergence curves, we can observe the following phenomena: 
(1) Compared with CSRNet, TransCrowd-GAP achieves better performance with 1.9 $\times$ fewer training epochs.
(2) Using global average pooled visual tokens can converge faster and achieve better count accuracy than using the extra regression token. (3) Both TransCrowd-Token and TransCrowd-GAP present a smooth curve and fast converging, while the curve of CSRNet is oscillating.
These observations show the potential value of the Transformer in the counting task.

\subsection{Comparison of run-time.}

As shown in Tab.~\ref{tab:runtime}, we compare with two popular fully supervised counting methods, including BL~\cite{ma2019bayesian} and CSRNET~\cite{li2018csrnet}. The experiment is conducted on a Titan XP GPU. Even though both the proposed TransCrowd-Token and TransCrowd-GAP contain more parameters than other methods, they still achieve outstanding run-time. This is because the fully supervised methods need to maintain high-resolution features to generate high-quality density maps (e.g., 1/8 size of the input in CSRNET~\cite{li2018csrnet} and 1/16 size of the input in BL~\cite{ma2019bayesian}). Additionally, we can observe that the FPS of VGG19-based BL~\cite{ma2019bayesian} outperforms the VGG16-based CSRNet~\cite{li2018csrnet}, mainly because the BL generates a small-resolution density map (1/16 of the input image). This phenomenon further demonstrates the influence of feature resolution on run-time.

\subsection{Comparison of Different Pre-trained strategies.} 
In this section, we study the impact of the pre-trained model in TransCrowd. We choose the popular CNN-based method CSRNet~\cite{li2018csrnet} as a comparison, and the results are listed in Tab.~\ref{tab:pretrained}.
Specifically, there are three strategies: (1) \textbf{None}: The models are directly trained on ShanghaiTech part A. (2) \textbf{Pre-ImgNet}: The models are pre-trained on the ImageNet and fine-tune on ShanghaiTech part A. (3) \textbf{Pre-GCC}: The models are  pre-trained on GCC~\cite{wang2019learning}, a synthetic dataset, and are fine-tuned on ShangahiTech part A dataset.

From the Tab.~\ref{tab:pretrained}, there are some interesting findings:
1) Without any pre-trained dataset, the CNN-based method outperforms the Transformer-based method. 2) Using the extra pre-trained data can effectively prompt the performance, and the proposed TransCrowd-GAP achieves better counting performance than CSRNet. 3) Besides, when the model pre-trained on the GCC dataset, the proposed method can even outperform several recent fully-supervised methods (e.g., CFF~\cite{shi2019counting}, TEDNet~\cite{jiang2019crowd}). Note that the GCC dataset is a synthetic crowd dataset, without any annotation cost, which means the TransCrowd-GAP can achieve similar counting performance to the fully-supervised methods by using small count-level labeled real-data and extensive free synthetic data, promoting the practical applications. It is noteworthy that the proposed method only uses count-level annotations of the GCC dataset, different from the previous fully-supervised work.

\subsection{Cross-dataset evaluation.} 
Finally, we conduct cross-dataset experiments on the UCF-QNRF, ShanghaiTech Part A and Part B datasets to explore the transferability of the proposed TransCrowd-GAP. In the Cross-dataset evaluation, models are trained on the source dataset and tested on the target dataset without further ﬁne-tuning. Quantitative results are shown in Tab.~\ref{tab:transfer}. Although our method is a weakly-supervised paradigm, we still achieve highly competitive performance, which shows remarkable transferability. 

\section{Conclusion}
In this work, we present an alternative perspective for weakly-supervised crowd counting in images by introducing a sequence-to-count prediction framework based on Transformer-encoder, named TransCrowd. To the best of our knowledge, we are the first to solve the counting problem based on the Transformer. We analyze and show that the attention mechanism is very promising to capture the semantic crowd information. Extensive experiments on five challenging datasets demonstrate that TransCrowd achieves superior counting performance compared with the state-of-the-art weakly-supervised methods and achieves competitive performance compared with some popular fully-supervised methods. In the future, we plan to make fully-supervised counting using Transformer architecture and extend it to video-based counting task.

\bibliographystyle{ieee_fullname}
\bibliography{egbib}

\begin{thebibliography}{10}\itemsep=-1pt

\bibitem{abousamra2020localization}
Shahira Abousamra, Minh Hoai, Dimitris Samaras, and Chao Chen.
\newblock Localization in the crowd with topological constraints.
\newblock In {\em Proc. of the AAAI Conf. on Artificial Intelligence}, 2021.

\bibitem{bai2020adaptive}
Shuai Bai, Zhiqun He, Yu Qiao, Hanzhe Hu, Wei Wu, and Junjie Yan.
\newblock Adaptive dilated network with self-correction supervision for
  counting.
\newblock In {\em Proc. of IEEE Intl. Conf. on Computer Vision and Pattern
  Recognition}, 2020.

\bibitem{cao2018scale}
Xinkun Cao, Zhipeng Wang, Yanyun Zhao, and Fei Su.
\newblock Scale aggregation network for accurate and efficient crowd counting.
\newblock In {\em Proc. of European Conference on Computer Vision}, 2018.

\bibitem{detr}
Nicolas Carion, Francisco Massa, Gabriel Synnaeve, Nicolas Usunier, Alexander
  Kirillov, and Sergey Zagoruyko.
\newblock End-to-end object detection with transformers.
\newblock In {\em Proc. of European Conference on Computer Vision}, 2020.

\bibitem{chan2008privacy}
Antoni~B Chan, Zhang-Sheng~John Liang, and Nuno Vasconcelos.
\newblock Privacy preserving crowd monitoring: Counting people without people
  models or tracking.
\newblock In {\em Proc. of IEEE Intl. Conf. on Computer Vision and Pattern
  Recognition}, 2008.

\bibitem{pretrained}
Hanting Chen, Yunhe Wang, Tianyu Guo, Chang Xu, Yiping Deng, Zhenhua Liu, Siwei
  Ma, Chunjing Xu, Chao Xu, and Wen Gao.
\newblock Pre-trained image processing transformer.
\newblock In {\em Proc. of IEEE Intl. Conf. on Computer Vision and Pattern
  Recognition}, pages 12299--12310, 2021.

\bibitem{chen2021cell}
Yajie Chen, Dingkang Liang, Xiang Bai, Yongchao Xu, and Xin Yang.
\newblock Cell localization and counting using direction field map.
\newblock {\em IEEE Journal of Biomedical and Health Informatics}, 2021.

\bibitem{du2020visdrone}
Dawei Du, Longyin Wen, Pengfei Zhu, Heng Fan, Qinghua Hu, Haibin Ling, Mubarak
  Shah, Junwen Pan, Ali Al-Ali, Amr Mohamed, et~al.
\newblock Visdrone-cc2020: The vision meets drone crowd counting challenge
  results.
\newblock In {\em Proc. of European Conference on Computer Vision}, pages
  675--691. Springer, 2020.

\bibitem{gao2019c}
Junyu Gao, Wei Lin, Bin Zhao, Dong Wang, Chenyu Gao, and Jun Wen.
\newblock C\^{} 3 framework: An open-source pytorch code for crowd counting.
\newblock {\em arXiv preprint arXiv:1907.02724}, 2019.

\bibitem{gao2019pcc}
Junyu Gao, Qi Wang, and Xuelong Li.
\newblock Pcc net: Perspective crowd counting via spatial convolutional
  network.
\newblock {\em IEEE Transactions on Circuits and Systems for Video Technology},
  2019.

\bibitem{guo2015mobile}
Bin Guo, Zhu Wang, Zhiwen Yu, Yu Wang, Neil~Y Yen, Runhe Huang, and Xingshe
  Zhou.
\newblock Mobile crowd sensing and computing: The review of an emerging
  human-powered sensing paradigm.
\newblock {\em ACM computing surveys (CSUR)}, 48(1):1--31, 2015.

\bibitem{hendrycks2016bridging}
Dan Hendrycks and Kevin Gimpel.
\newblock Bridging nonlinearities and stochastic regularizers with gaussian
  error linear units.
\newblock 2016.

\bibitem{idrees2013multi}
Haroon Idrees, Imran Saleemi, Cody Seibert, and Mubarak Shah.
\newblock Multi-source multi-scale counting in extremely dense crowd images.
\newblock In {\em Proc. of IEEE Intl. Conf. on Computer Vision and Pattern
  Recognition}, 2013.

\bibitem{idrees2018composition}
Haroon Idrees, Muhmmad Tayyab, Kishan Athrey, Dong Zhang, Somaya Al-Maadeed,
  Nasir Rajpoot, and Mubarak Shah.
\newblock Composition loss for counting, density map estimation and
  localization in dense crowds.
\newblock In {\em Proc. of European Conference on Computer Vision}, 2018.

\bibitem{jiang2019crowd}
Xiaolong Jiang, Zehao Xiao, Baochang Zhang, Xiantong Zhen, Xianbin Cao, David
  Doermann, and Ling Shao.
\newblock Crowd counting and density estimation by trellis encoder-decoder
  networks.
\newblock In {\em Proc. of IEEE Intl. Conf. on Computer Vision and Pattern
  Recognition}, 2019.

\bibitem{jiang2020attention}
Xiaoheng Jiang, Li Zhang, Mingliang Xu, Tianzhu Zhang, Pei Lv, Bing Zhou, Xin
  Yang, and Yanwei Pang.
\newblock Attention scaling for crowd counting.
\newblock In {\em Proc. of IEEE Intl. Conf. on Computer Vision and Pattern
  Recognition}, 2020.

\bibitem{devlin2018bert}
Jacob Devlin Ming-Wei~Chang Kenton and Lee~Kristina Toutanova.
\newblock Bert: Pre-training of deep bidirectional transformers for language
  understanding.
\newblock In {\em Proceedings of NAACL-HLT}, pages 4171--4186, 2019.

\bibitem{kingma2014adam}
DP Kingma and JL Ba.
\newblock Adam: A method for stochastic optimization 3rd international
  conference on learning representations.
\newblock In {\em Proc. of International Conference on Learning
  Representations}, 2015.

\bibitem{vit}
Alexander Kolesnikov, Alexey Dosovitskiy, Dirk Weissenborn, Georg Heigold,
  Jakob Uszkoreit, Lucas Beyer, Matthias Minderer, Mostafa Dehghani, Neil
  Houlsby, Sylvain Gelly, et~al.
\newblock An image is worth 16x16 words: Transformers for image recognition at
  scale.
\newblock 2021.

\bibitem{lei2021towards}
Yinjie Lei, Yan Liu, Pingping Zhang, and Lingqiao Liu.
\newblock Towards using count-level weak supervision for crowd counting.
\newblock {\em Pattern Recognition}, 109:107616, 2021.

\bibitem{li2018csrnet}
Yuhong Li, Xiaofan Zhang, and Deming Chen.
\newblock {CSRNet}: Dilated convolutional neural networks for understanding the
  highly congested scenes.
\newblock In {\em Proc. of IEEE Intl. Conf. on Computer Vision and Pattern
  Recognition}, 2018.

\bibitem{liang2021focal}
Dingkang Liang, Wei Xu, Yingying Zhu, and Yu Zhou.
\newblock Focal inverse distance transform maps for crowd localization and
  counting in dense crowd.
\newblock {\em arXiv preprint arXiv:2102.07925}, 2021.

\bibitem{lin2021direct}
Hui Lin, Xiaopeng Hong, Zhiheng Ma, Xing Wei, Yunfeng Qiu, Yaowei Wang, and
  Yihong Gong.
\newblock Direct measure matching for crowd counting.
\newblock In {\em Proceedings of the Thirtieth International Joint Conference
  on Artificial Intelligence}, 2021.

\bibitem{liu2018decidenet}
Jiang Liu, Chenqiang Gao, Deyu Meng, and Alexander~G Hauptmann.
\newblock Decidenet: counting varying density crowds through attention guided
  detection and density estimation.
\newblock In {\em Proc. of IEEE Intl. Conf. on Computer Vision and Pattern
  Recognition}, 2018.

\bibitem{liu2020weighing}
Liang Liu, Hao Lu, Hongwei Zou, Haipeng Xiong, Zhiguo Cao, and Chunhua Shen.
\newblock Weighing counts: Sequential crowd counting by reinforcement learning.
\newblock In {\em Proc. of European Conference on Computer Vision}, pages
  164--181. Springer, 2020.

\bibitem{liu2019crowd}
Lingbo Liu, Zhilin Qiu, Guanbin Li, Shufan Liu, Wanli Ouyang, and Liang Lin.
\newblock Crowd counting with deep structured scale integration network.
\newblock In {\em Porc. of IEEE Intl. Conf. on Computer Vision}, 2019.

\bibitem{liu2019adcrowdnet}
Ning Liu, Yongchao Long, Changqing Zou, Qun Niu, Li Pan, and Hefeng Wu.
\newblock Adcrowdnet: An attention-injective deformable convolutional network
  for crowd understanding.
\newblock In {\em Proc. of IEEE Intl. Conf. on Computer Vision and Pattern
  Recognition}, 2019.

\bibitem{liu2016ssd}
Wei Liu, Dragomir Anguelov, Dumitru Erhan, Christian Szegedy, Scott Reed,
  Cheng-Yang Fu, and Alexander~C Berg.
\newblock Ssd: Single shot multibox detector.
\newblock In {\em Proc. of European Conference on Computer Vision}, 2016.

\bibitem{liu2019context}
Weizhe Liu, Mathieu Salzmann, and Pascal Fua.
\newblock Context-aware crowd counting.
\newblock In {\em Proc. of IEEE Intl. Conf. on Computer Vision and Pattern
  Recognition}, 2019.

\bibitem{liu2019exploiting}
Xialei Liu, Joost Van De~Weijer, and Andrew~D Bagdanov.
\newblock Exploiting unlabeled data in cnns by self-supervised learning to
  rank.
\newblock {\em IEEE transactions on pattern analysis and machine intelligence},
  2019.

\bibitem{liu2019roberta}
Yinhan Liu, Myle Ott, Naman Goyal, Jingfei Du, Mandar Joshi, Danqi Chen, Omer
  Levy, Mike Lewis, Luke Zettlemoyer, and Veselin Stoyanov.
\newblock Roberta: A robustly optimized bert pretraining approach.
\newblock {\em arXiv preprint arXiv:1907.11692}, 2019.

\bibitem{liu2019point}
Yuting Liu, Miaojing Shi, Qijun Zhao, and Xiaofang Wang.
\newblock Point in, box out: Beyond counting persons in crowds.
\newblock In {\em Proc. of IEEE Intl. Conf. on Computer Vision and Pattern
  Recognition}, 2019.

\bibitem{liu2021visdrone}
Zhihao Liu, Zhijian He, Lujia Wang, Wenguan Wang, Yixuan Yuan, Dingwen Zhang,
  Jinglin Zhang, Pengfei Zhu, Luc Van~Gool, Junwei Han, et~al.
\newblock Visdrone-cc2021: The vision meets drone crowd counting challenge
  results.
\newblock In {\em Porc. of IEEE Intl. Conf. on Computer Vision}, pages
  2830--2838, 2021.

\bibitem{ma2019bayesian}
Zhiheng Ma, Xing Wei, Xiaopeng Hong, and Yihong Gong.
\newblock Bayesian loss for crowd count estimation with point supervision.
\newblock In {\em Porc. of IEEE Intl. Conf. on Computer Vision}, 2019.

\bibitem{ma2020learning}
Zhiheng Ma, Xing Wei, Xiaopeng Hong, and Yihong Gong.
\newblock Learning scales from points: A scale-aware probabilistic model for
  crowd counting.
\newblock In {\em Proc. of ACM Multimedia}, pages 220--228, 2020.

\bibitem{ma2021learning}
Zhiheng Ma, Xing Wei, Xiaopeng Hong, Hui Lin, Yunfeng Qiu, and Yihong Gong.
\newblock Learning to count via unbalanced optimal transport.
\newblock In {\em Proc. of the AAAI Conf. on Artificial Intelligence},
  volume~35, pages 2319--2327, 2021.

\bibitem{ranjan2018iterative}
Viresh Ranjan, Hieu Le, and Minh Hoai.
\newblock Iterative crowd counting.
\newblock In {\em Proc. of European Conference on Computer Vision}, 2018.

\bibitem{ren2015faster}
Shaoqing Ren, Kaiming He, Ross Girshick, and Jian Sun.
\newblock Faster r-cnn: Towards real-time object detection with region proposal
  networks.
\newblock In {\em Proc. of Advances in Neural Information Processing Systems},
  2015.

\bibitem{sam2020locate}
Deepak~Babu Sam, Skand~Vishwanath Peri, Mukuntha~Narayanan Sundararaman, Amogh
  Kamath, and Venkatesh~Babu Radhakrishnan.
\newblock Locate, size and count: Accurately resolving people in dense crowds
  via detection.
\newblock {\em IEEE Transactions on Pattern Analysis and Machine Intelligence},
  2020.

\bibitem{sheng2014leveraging}
Xiang Sheng, Jian Tang, Xuejie Xiao, and Guoliang Xue.
\newblock Leveraging gps-less sensing scheduling for green mobile crowd
  sensing.
\newblock {\em IEEE Internet of Things Journal}, 1(4):328--336, 2014.

\bibitem{shi2019revisiting}
Miaojing Shi, Zhaohui Yang, Chao Xu, and Qijun Chen.
\newblock Revisiting perspective information for efficient crowd counting.
\newblock In {\em Proc. of IEEE Intl. Conf. on Computer Vision and Pattern
  Recognition}, 2019.

\bibitem{shi2019counting}
Zenglin Shi, Pascal Mettes, and Cees~GM Snoek.
\newblock Counting with focus for free.
\newblock In {\em Porc. of IEEE Intl. Conf. on Computer Vision}, pages
  4200--4209, 2019.

\bibitem{shi2018crowd}
Zenglin Shi, Le Zhang, Yun Liu, Xiaofeng Cao, Yangdong Ye, Ming-Ming Cheng, and
  Guoyan Zheng.
\newblock Crowd counting with deep negative correlation learning.
\newblock In {\em Proc. of IEEE Intl. Conf. on Computer Vision and Pattern
  Recognition}, 2018.

\bibitem{sindagi2020jhu}
Vishwanath Sindagi, Rajeev Yasarla, and Vishal~MM Patel.
\newblock Jhu-crowd++: Large-scale crowd counting dataset and a benchmark
  method.
\newblock {\em IEEE Transactions on Pattern Analysis and Machine Intelligence},
  2020.

\bibitem{sindagi2017cnn}
Vishwanath~A Sindagi and Vishal~M Patel.
\newblock Cnn-based cascaded multi-task learning of high-level prior and
  density estimation for crowd counting.
\newblock In {\em Proc. of IEEE Intl. Conf. on Advanced Video and Signal Based
  Surveillance}, 2017.

\bibitem{sindagi2017generating}
Vishwanath~A Sindagi and Vishal~M Patel.
\newblock Generating high-quality crowd density maps using contextual pyramid
  cnns.
\newblock In {\em Porc. of IEEE Intl. Conf. on Computer Vision}, 2017.

\bibitem{sindagi2019multi}
Vishwanath~A Sindagi and Vishal~M Patel.
\newblock Multi-level bottom-top and top-bottom feature fusion for crowd
  counting.
\newblock In {\em Porc. of IEEE Intl. Conf. on Computer Vision}, 2019.

\bibitem{vaswani2017attention}
Ashish Vaswani, Noam Shazeer, Niki Parmar, Jakob Uszkoreit, Llion Jones,
  Aidan~N Gomez, Lukasz Kaiser, and Illia Polosukhin.
\newblock Attention is all you need.
\newblock In {\em Proc. of Advances in Neural Information Processing Systems},
  2017.

\bibitem{von2016gaussian}
Matthias von Borstel, Melih Kandemir, Philip Schmidt, Madhavi~K Rao, Kumar
  Rajamani, and Fred~A Hamprecht.
\newblock Gaussian process density counting from weak supervision.
\newblock In {\em Proc. of European Conference on Computer Vision}, pages
  365--380. Springer, 2016.

\bibitem{wan2020modeling}
Jia Wan and Antoni Chan.
\newblock Modeling noisy annotations for crowd counting.
\newblock {\em Advances in Neural Information Processing Systems}, 2020.

\bibitem{wan2019residual}
Jia Wan, Wenhan Luo, Baoyuan Wu, Antoni~B Chan, and Wei Liu.
\newblock Residual regression with semantic prior for crowd counting.
\newblock In {\em Proc. of IEEE Intl. Conf. on Computer Vision and Pattern
  Recognition}, 2019.

\bibitem{wan2020kernel}
Jia Wan, Qingzhong Wang, and Antoni~B Chan.
\newblock Kernel-based density map generation for dense object counting.
\newblock {\em IEEE Transactions on Pattern Analysis and Machine Intelligence},
  2020.

\bibitem{wang2020distribution}
Boyu Wang, Huidong Liu, Dimitris Samaras, and Minh Hoai.
\newblock Distribution matching for crowd counting.
\newblock In {\em Proc. of Advances in Neural Information Processing Systems},
  2020.

\bibitem{transmeettrack}
Ning Wang, Wengang Zhou, Jie Wang, and Houqiang Li.
\newblock Transformer meets tracker: Exploiting temporal context for robust
  visual tracking.
\newblock In {\em Proc. of IEEE Intl. Conf. on Computer Vision and Pattern
  Recognition}, pages 1571--1580, 2021.

\bibitem{gao2020nwpu}
Qi Wang, Junyu Gao, Wei Lin, and Xuelong Li.
\newblock Nwpu-crowd: A large-scale benchmark for crowd counting and
  localization.
\newblock {\em IEEE Transactions on Pattern Analysis and Machine Intelligence},
  2020.

\bibitem{wang2019learning}
Qi Wang, Junyu Gao, Wei Lin, and Yuan Yuan.
\newblock Learning from synthetic data for crowd counting in the wild.
\newblock In {\em Proc. of IEEE Intl. Conf. on Computer Vision and Pattern
  Recognition}, 2019.

\bibitem{xu2019autoscale}
Chenfeng Xu, Dingkang Liang, Yongchao Xu, Song Bai, Wei Zhan, Xiang Bai, and
  Masayoshi Tomizuka.
\newblock Autoscale: Learning to scale for crowd counting.
\newblock {\em International Journal of Computer Vision}, pages 1--30, 2022.

\bibitem{xu2019learn}
Chenfeng Xu, Kai Qiu, Jianlong Fu, Song Bai, Yongchao Xu, and Xiang Bai.
\newblock Learn to scale: Generating multipolar normalized density map for
  crowd counting.
\newblock In {\em Porc. of IEEE Intl. Conf. on Computer Vision}, 2019.

\bibitem{xu2020dilated}
Wei Xu, Dingkang Liang, Yixiao Zheng, Jiahao Xie, and Zhanyu Ma.
\newblock Dilated-scale-aware category-attention convnet for multi-class object
  counting.
\newblock {\em IEEE Signal Processing Letters}, 28:1570--1574, 2021.

\bibitem{yan2019perspective}
Zhaoyi Yan, Yuchen Yuan, Wangmeng Zuo, Xiao Tan, Yezhen Wang, Shilei Wen, and
  Errui Ding.
\newblock Perspective-guided convolution networks for crowd counting.
\newblock In {\em Porc. of IEEE Intl. Conf. on Computer Vision}, 2019.

\bibitem{yang2020reverse}
Yifan Yang, Guorong Li, Zhe Wu, Li Su, Qingming Huang, and Nicu Sebe.
\newblock Reverse perspective network for perspective-aware object counting.
\newblock In {\em Proc. of IEEE Intl. Conf. on Computer Vision and Pattern
  Recognition}, 2020.

\bibitem{yang2020weakly}
Yifan Yang, Guorong Li, Zhe Wu, Li Su, Qingming Huang, and Nicu Sebe.
\newblock Weakly-supervised crowd counting learns from sorting rather than
  locations.
\newblock In {\em Proc. of European Conference on Computer Vision}, 2020.

\bibitem{zhang2019attentional}
Anran Zhang, Lei Yue, Jiayi Shen, Fan Zhu, Xiantong Zhen, Xianbin Cao, and Ling
  Shao.
\newblock Attentional neural fields for crowd counting.
\newblock In {\em Porc. of IEEE Intl. Conf. on Computer Vision}, 2019.

\bibitem{zhang2015cross}
Cong Zhang, Hongsheng Li, Xiaogang Wang, and Xiaokang Yang.
\newblock Cross-scene crowd counting via deep convolutional neural networks.
\newblock In {\em Proc. of IEEE Intl. Conf. on Computer Vision and Pattern
  Recognition}, pages 833--841, 2015.

\bibitem{zhang2017fcn}
Shanghang Zhang, Guanhang Wu, Joao~P Costeira, and Jos{\'e}~MF Moura.
\newblock Fcn-rlstm: Deep spatio-temporal neural networks for vehicle counting
  in city cameras.
\newblock In {\em Porc. of IEEE Intl. Conf. on Computer Vision}, 2017.

\bibitem{zhang2016single}
Yingying Zhang, Desen Zhou, Siqin Chen, Shenghua Gao, and Yi Ma.
\newblock Single-image crowd counting via multi-column convolutional neural
  network.
\newblock In {\em Proc. of IEEE Intl. Conf. on Computer Vision and Pattern
  Recognition}, 2016.

\bibitem{SETR}
Sixiao Zheng, Jiachen Lu, Hengshuang Zhao, Xiatian Zhu, Zekun Luo, Yabiao Wang,
  Yanwei Fu, Jianfeng Feng, Tao Xiang, Philip~HS Torr, et~al.
\newblock Rethinking semantic segmentation from a sequence-to-sequence
  perspective with transformers.
\newblock In {\em Proc. of IEEE Intl. Conf. on Computer Vision and Pattern
  Recognition}, pages 6881--6890, 2021.

\bibitem{deformable_detr}
Xizhou Zhu, Weijie Su, Lewei Lu, Bin Li, Xiaogang Wang, and Jifeng Dai.
\newblock Deformable detr: Deformable transformers for end-to-end object
  detection.
\newblock In {\em International Conference on Learning Representations}, 2020.

\end{thebibliography}

\end{document}